%% file: fnf_arxiv.tex
\documentclass{iopjournal}
\usepackage[utf8]{inputenc} 
\usepackage[T1]{fontenc}    

\usepackage{hyperref}       
\usepackage{url}            
\usepackage{booktabs}       
\usepackage{amsfonts}       
\usepackage{nicefrac}       
\usepackage{microtype}      
\usepackage{lipsum}		
\usepackage{graphicx}
\usepackage{subcaption}
\usepackage[numbers,sort&compress]{natbib}
\usepackage{amsmath}
\usepackage{doi}
\usepackage{tikz}
\usepackage{standalone}

\usetikzlibrary{positioning,calc,fit,backgrounds,arrows.meta}

\renewcommand{\articletype}[1]{{\vspace*{-8mm}\noindent}

\vspace*{8mm} \noindent {\scriptsize \sf{\bfseries \MakeUppercase{#1}}}}

\begin{document}

\articletype{Paper} 

\title{Factorizable Normalizing Flows for parameter-dependent density morphing}

\author{Davide Valsecchi$^1$\orcid{0000-0001-8587-8266}}
\author{Mauro Doneg\`a$^1$\orcid{0000-0001-9830-0412}}
\author{Rainer Wallny$^1$\orcid{0000-0001-8038-1613}}

\affil{$^1$Institute for Particle Physics and Astrophysics, D-PHYS, ETH Zurich, Zurich, Switzerland}

\email{dvalsecchi@ethz.ch, mdonega@ethz.ch, rwallny@ethz.ch}


\begin{abstract}
Normalizing Flows excel at modeling a single fixed density, yet many problems across the sciences, such as high energy physics, instead require modeling how that density \emph{deforms} as a function of continuous parameters: the strength of a physical effect, a calibration constant, or a source of systematic uncertainty. Learning a separate flow for every parameter configuration quickly becomes intractable, since the number of joint settings grows exponentially with the number of parameters. We introduce \textbf{Factorizable Normalizing Flows} (FNFs), which represent the parameter-dependent density as a fixed, high-fidelity flow for a reference configuration composed with a learnable transformation that is polynomial in the parameters and factorized over them. This structure has a practical consequence: each parameter's effect is learned in isolation, from samples in which that parameter alone is varied. The combined response of many parameters is then recovered by summation at inference, without ever sampling their combinatorially large joint space. On a controlled problem with two interpretable deformations applied jointly to the data, the learned transformation reproduces the true deformations and matches the optimal likelihood, while optional interaction terms capture residual correlations when several parameters vary strongly at once. The resulting model is interpretable, scales linearly with the number of parameters, and keeps the likelihood tractable. This provides a general tool for any inference workflow requiring continuous density morphing, and directly enables the next generation of unbinned likelihood fits in high energy physics.
\end{abstract}


\section{Introduction}
Estimating probability density functions (PDFs) from data is a central task across the quantitative sciences and the backbone of likelihood-based and simulation-based inference (SBI)~\cite{Cranmer:2020wdu}, where parameters of interest are extracted by comparing observed data to a model of its expected distribution. When the PDF cannot be written in closed form, or is accessible only through an expensive forward simulation, it must be learned from samples.

Traditionally such densities are estimated with histograms or kernel density estimators, which struggle in high dimensions. Normalizing Flows (NFs)~\cite{Kobyzev:2020nf} have emerged as a powerful alternative: they learn an invertible mapping between a simple base distribution, such as a multivariate Gaussian, and the target density, giving both efficient sampling and an explicit, normalized likelihood in high-dimensional spaces. They have found widespread success across the physical sciences, with applications ranging from astrophysics to lattice field theory~\cite{p3f7-rbmv,Papamakarios:2021nf,Noe:2019boltzmann,Albergo:2019eim}.

In many settings, however, a single fixed density is not enough: the object of interest is how the density \emph{changes} as a function of continuous parameters. These may be physical parameters of interest, instrument calibrations, or systematic uncertainties that reshape the distribution. This dependence is rarely known analytically, so it too must be inferred from simulations produced at different parameter values. Doing so naively does not scale: capturing the joint effect of $K$ parameters would require sampling their full $K$-dimensional configuration space, whose cost grows exponentially with $K$.

This challenge is acute in high energy physics. At the LHC, parameters of interest are extracted by fitting observed data against simulated templates while \emph{profiling} hundreds of nuisance parameters that encode systematic uncertainties~\cite{CMS:2024combine,Cranmer:2012sba}. The effect of each uncertainty is propagated through a pair of $\pm 1\sigma$ template variations and interpolated with a low-order polynomial: a robust and widely used procedure, but one tied to binned, low-dimensional summaries of the data. Unbinned fits over high-dimensional observables would extract substantially more information per event, provided the response of the \emph{full} density to each systematic can be modeled accurately and propagated to the likelihood at comparable scale.

In this work we introduce \textbf{Factorizable Normalizing Flows} (FNFs), a scalable approach to this parametric density-morphing problem. An FNF composes a fixed, high-fidelity flow for a reference configuration with a learnable transformation that is polynomial in the parameters and factorized over them, so that the effect of each parameter is learned in isolation and the joint response is recovered by summation at inference. This directly generalizes the $\pm 1\sigma$ template-variation procedure to a continuous, differentiable, high-dimensional setting, while requiring only the per-systematic variations that analyses already produce: the combined nuisance space is never sampled. Here we focus on validating the parametric density modeling itself: we show that the FNF reproduces the imposed deformations and closely matches the optimal likelihood, with optional interaction terms capturing the residual correlations between parameters. Its deployment in a profiled, unbinned likelihood fit to extract parameters of interest is the subject of dedicated forthcoming work.

This paper is organized as follows: in Section~\ref{sec:systematics} we discuss the challenges of modeling systematic uncertainties in unbinned likelihood fits and motivate the need for a new approach. In Section~\ref{sec:factorizable_flows} we introduce the Factorizable Normalizing Flow architecture and describe its key components. In Sections~\ref{sec:experiments} and~\ref{sec:discussion} we present experimental results demonstrating the effectiveness of our approach on a toy dataset, and in Section~\ref{sec:related_work} we discuss related work in the field. Finally, we conclude with a summary of our findings and potential future directions in Section~\ref{sec:conclusions}.

\begin{figure}[t]
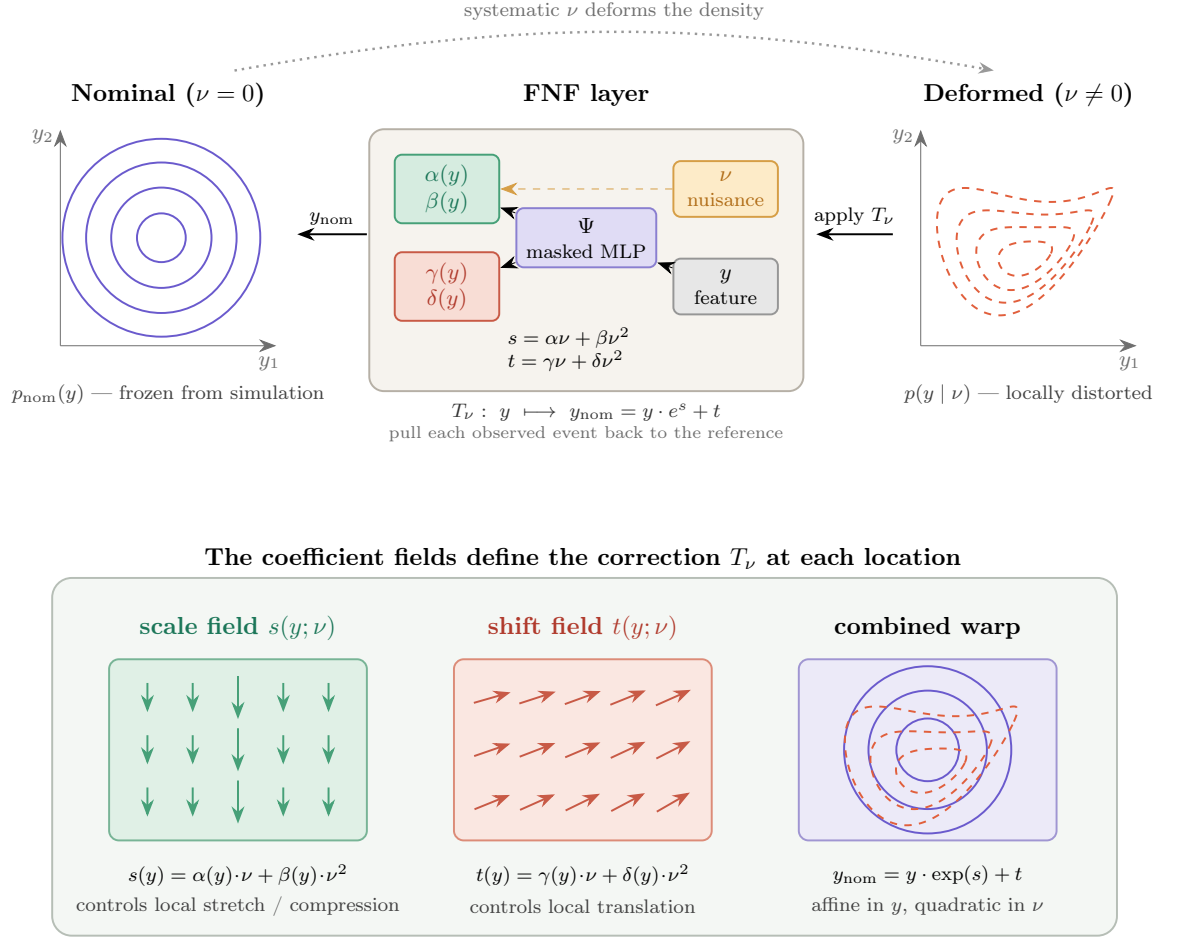

  \centering
  \includestandalone[width=\textwidth]{figs/fnf_schema_tikz}
  \caption{Schematic of the Factorizable Normalizing Flow. A systematic
  $\nu$ deforms the nominal reference density $p_{\mathrm{nom}}(y)$ into the
  observed density $p(y\mid\nu)$ (dotted arrow, top). The FNF models this
  effect through its \emph{inverse}: a learnable transformation $T_\nu$ that
  pulls each observed event back to the reference,
  $y_{\mathrm{nom}}=T_\nu(y)=y\,e^{s}+t$, where it is scored under the frozen
  nominal flow. The scale $s$ and shift $t$ are built from coefficient fields
  output by a masked MLP $\Psi$ and are polynomial in $\nu$ (bottom).}
  \label{fig:fnf_schema}
\end{figure}

\section{Systematic uncertainties in unbinned likelihood fits}
\label{sec:systematics}
We consider a typical high energy physics (HEP) measurement, where the goal is to infer parameters of interest from observed data while accounting for systematic uncertainties. Each source of uncertainty is encoded by a nuisance parameter, collected in a vector $\nu$, and the central object of the analysis is the conditional density $p(y \mid x, \nu)$, where $y$ are the observed features whose distribution we model, $x$ are conditioning variables, and $\nu$ are the nuisance parameters. Traditionally the likelihood is built by \emph{binning} the data in one or two features and comparing the observed and expected counts per bin; an unbinned fit in higher dimensions instead requires the full density $p(y \mid x, \nu)$ itself. In either case the dependence on $\nu$ is not known analytically and must be learned from simulation, which is the central modeling challenge addressed in this work.

In HEP, the effect of systematic uncertainties is traditionally modeled through \textbf{template variations}~\cite{CMS:2024combine,Cranmer:2012sba}. The impact of each uncertainty source is captured \emph{independently}: the corresponding nuisance is shifted to the edges of its $68\%$ confidence interval ($\nu_k = \pm 1$, the ``$\pm 1\sigma$'' variations), the simulation is re-run, and the resulting ``up'' and ``down'' histograms bracket the nominal one. The per-bin yield is then interpolated as a smooth polynomial in each nuisance, of at least quadratic order~\cite{Baak:2014fta}, so that the nuisances can be \emph{profiled} continuously during the likelihood fit~\cite{Lista:2023stat}. This construction is the workhorse of LHC analyses: it is robust, and because each source is treated separately it scales to hundreds of nuisance parameters. Its limitations are equally structural: it operates on binned, low-dimensional summaries of the data, and it assumes that the effects of different sources \emph{factorize}, neglecting their correlations, an approximation that simplifies the fit but can bias it.

Carrying this idea to unbinned, high-dimensional fits places several demands on the modeling method:
\begin{itemize}
  \item \textbf{Fidelity}: it must accurately capture the effect of each systematic on the input density, so that the learned deformation reflects the true response to each source.
  \item \textbf{Profiling}: it must expose a smooth, differentiable dependence on $\nu$, so that the nuisances can be profiled during the fit.
  \item \textbf{Scalability}: it must scale gracefully with the number of nuisances, both in the simulation required to train it (no sampling of the joint nuisance space) and in the cost of evaluating the density and its Jacobian during the fit.
  \item \textbf{Interpretability}: the learned transformation should remain interpretable, exposing how each nuisance reshapes the distribution.
\end{itemize}

A direct way to bring this picture to generative models is to model the effect of systematic uncertainties as \textbf{parametric deformations} of a learned density. Two routes are available:
\begin{enumerate}
  \item make the generative model directly conditional on the nuisances, learning $p(y \mid x, \nu)$ end-to-end from samples generated at different values of $\nu$;
  \item keep a fixed nominal density and compose it with an additional transformation that captures the deformation induced by the nuisances, $p(y \mid x, \nu) = p_{\text{nom}}(y \mid x) \circ T(y \mid x, \nu)$.
\end{enumerate}

The first route is conceptually simple but does not scale: to capture the joint effect of $K$ nuisances the model must be trained on samples spanning their full $K$-dimensional space, whose size grows exponentially with $K$. The second route is more promising, as it isolates the deformation in a dedicated transformation $T$ and, in the spirit of the template-variation approach, opens the door to \emph{factorizing} it over the individual nuisances. This is the route we take. Realizing it, however, requires a transformation that respects this factorization, stays polynomial in $\nu$ for smooth profiling, and keeps the likelihood tractable, a combination that calls for a dedicated architecture, which we introduce next.

\section{Factorizable Normalizing Flows}
\label{sec:factorizable_flows}
To address the challenge of scalable density estimation under systematic uncertainties, we introduce the \textbf{Factorizable Normalizing Flow} (FNF). Its central design choice is to structurally decouple the modeling of the complex, multidimensional distribution from the modeling of systematic variations: the density is a composition of a fixed, high-fidelity \emph{nominal model} $p_{\text{nom}}(y \mid x)$, trained once on high-statistics nominal simulation, and a learnable \emph{systematic transformation} $T_\nu$ that captures how probability mass migrates under systematic perturbations.

\subsection{The idea}
The FNF rests on a simple separation of concerns. Rather than learning, in one shot, both the shape of the data and how every systematic distorts it, we split the problem in two. First, we learn an accurate model of the data in its \emph{reference} configuration, the nominal simulation at $\nu = 0$, once and for all. Then, for each systematic, we learn a smooth \emph{deformation} of the feature space, a warping that morphs the reference distribution into the one observed when that systematic is switched on (Figure~\ref{fig:fnf_schema}).

To judge how likely an observed event is for a given setting of the systematics, we apply the deformation \emph{in reverse}: we carry the event back to the reference frame and ask how probable it was there, correcting for how much the warp stretched or compressed the space around it. This last correction is the familiar change-of-variables (Jacobian) factor that makes the construction a proper, normalized probability density.

The key property is that each systematic contributes its \emph{own} deformation and, to a good approximation, these deformations \emph{add up}. We can therefore learn each one in isolation, from just the $\pm 1\sigma$ samples already produced for that systematic, and sum their contributions to recover the joint response, with no need for simulations in which several systematics are varied together. This additivity is exact when the systematics act independently and otherwise an approximation, whose neglected correlations, and the optional interaction terms that restore them, we examine in Section~\ref{sec:additivity}. In this sense the FNF is a direct, continuous generalization of the template-variation recipe of Section~\ref{sec:systematics}: the discrete $\pm 1\sigma$ histograms become a smooth, learnable warp of the full density, and the per-bin polynomial interpolation becomes a polynomial dependence of that warp on $\nu$.

\subsection{Morphing as a pullback}
Mathematically, we treat the systematic-aware density as the \emph{pullback} of the nominal density by the diffeomorphism $T_\nu$. If $y$ is an observable in the ``distorted'' space (for example, data affected by detector effects), the transformation $T_\nu(y \mid x, \nu)$ maps it back to the ``nominal'' reference space $y_{\text{nom}}$. The likelihood is evaluated by correcting the event and scoring it under the nominal model, penalized by the volume change:
\begin{equation}
 y_{\text{nom}} = T_\nu(y \mid x, \nu) \quad \Rightarrow \quad p(y \mid x, \nu) = p_{\text{nom}}\!\left(T_\nu(y \mid x, \nu) \mid x\right) \left|\det \nabla_y T_\nu(y \mid x, \nu)\right|.
\end{equation}
For this construction to serve an unbinned fit, $T_\nu$ must meet the two requirements: its Jacobian determinant must be cheap to evaluate, and its dependence on $\nu$ must be polynomial and factorized over the individual nuisances. We satisfy both with the construction below.

\subsection{A tractable, factorized transformation}
We realize $T_\nu$ as an \emph{affine}~\cite{Dinh:2016realnvp}, \emph{autoregressive}~\cite{Kingma:2016iaf,Papamakarios:2017maf,Kobyzev:2020nf} transformation. The autoregressive structure makes the Jacobian triangular, so its determinant is simply the product of the diagonal and costs only $O(D)$ for $D$ feature dimensions. For each dimension $j$,
\begin{equation}
y_{\text{nom}, j} = y_j \cdot \exp\!\left(s_j(y_{<j}, x, \nu)\right) + t_j(y_{<j}, x, \nu),
\end{equation}
where the scale $s_j$ and shift $t_j$ depend only on the preceding features $y_{<j}$, the conditioning $x$, and the nuisances $\nu$.

\begin{figure}[htbp]
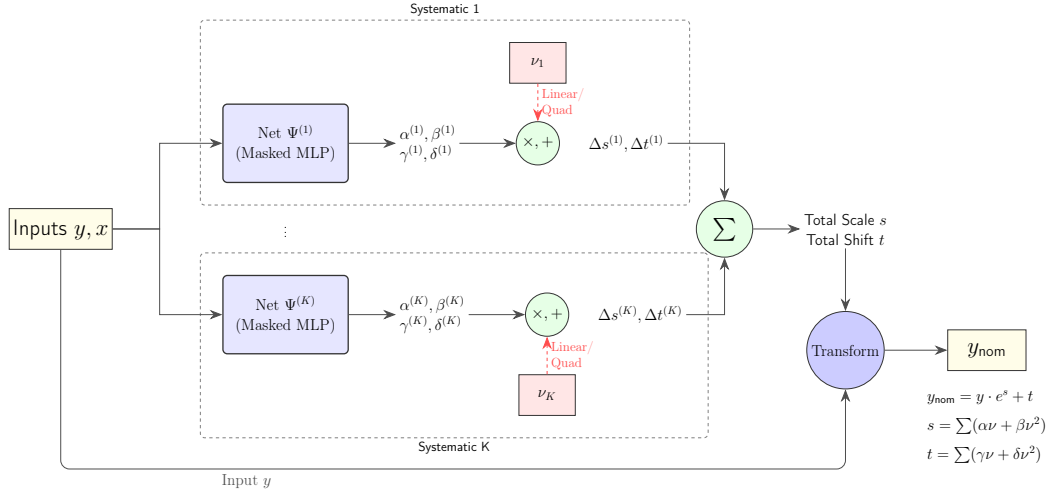

    \centering
    \includestandalone[width=0.9\textwidth]{figs/fnf_factorizable_tikz}
    \caption{Schematic representation of the Factorizable Normalizing Flow layer. The scale $s$ and shift $t$ parameters are computed as a sum of independent contributions from each systematic uncertainty $\nu_k$. Each contribution is parameterized as a quadratic function of $\nu_k$, with coefficients learned by a dedicated Masked MLP $\Psi^{k}$ conditioned on the inputs.}
    \label{fig:factorizable_flow}
\end{figure}

We interpret the systematic deformation as a \emph{Taylor expansion} of the transformation around the nominal point $\nu = 0$, and accordingly decompose the scale and shift as polynomials in $\nu$, factorized over the individual nuisances and including optional pairwise interaction terms (Figure~\ref{fig:factorizable_flow}):
\begin{align}\label{eq:factorizable_flows}
s_j(y_{<j}, x, \nu) &= \sum_{k=1}^K \left( \nu_k\,\alpha_{j}^{k}(y_{<j}, x) + \nu_k^2\,\beta_{j}^{k}(y_{<j}, x) \right) + \sum_{1 \le k < \ell \le K} \nu_k\,\nu_{\ell}\,\phi_{j}^{k\ell}(y_{<j}, x) \\
t_j(y_{<j}, x, \nu) &= \sum_{k=1}^K \left( \nu_k\,\gamma_{j}^{k}(y_{<j}, x) + \nu_k^2\,\delta_{j}^{k}(y_{<j}, x) \right) + \sum_{1 \le k < \ell \le K} \nu_k\,\nu_{\ell}\,\psi_{j}^{k\ell}(y_{<j}, x)
\end{align}
The dependence on $\nu$ is thus additive across nuisances, each acting through its own set of coefficients. This additivity in the affine parameters is what averts the curse of dimensionality of learning the joint conditional $p(y \mid x, \nu)$ directly, and what lets each nuisance be trained in isolation. The coefficients $\alpha, \beta, \gamma, \delta, \phi, \psi$ themselves carry the full kinematic complexity: they are the outputs of deep neural networks $\Psi^{k}$ conditioned on $y$ and $x$. We implement $\Psi^{k}$ as \textbf{masked multi-layer perceptrons} (masked MLPs)~\cite{Germain:2015made,Kobyzev:2020nf}, so a single network produces the autoregressive coefficients for all dimensions $j$ in one forward pass while respecting the causal constraint (dependence only on $y_{<j}$). The result is expressive enough to capture deep feature correlations yet structured enough to scale linearly with the number of systematics.

The cross-terms $\phi$ and $\psi$ are optional: they capture pairwise interactions between nuisances, but can be dropped whenever the systematics are treated as independent (factorizable), recovering the standard assumption of the template approach. Finally, stacking several such layers, interleaved with permutations~\cite{Kingma:2018glow} that mix the feature dimensions, yields a highly expressive transformation that retains precise, polynomial control over the dependence on $\nu$.

Together, these choices realize the transformation anticipated in Section~\ref{sec:systematics}: factorized over the nuisances, polynomial in $\nu$, and equipped with a tractable likelihood. The FNF thereby generalizes the template-variation method to a continuous, high-dimensional setting and, because the response is additive, each systematic can be trained independently from only its $\pm 1\sigma$ samples, with the joint effect recovered by summation. This is what makes the approach feasible even for analyses with many systematic uncertainties.

\section{Experiments}
\label{sec:experiments}
We validate the FNF on a controlled toy dataset that reproduces, in a low-dimensional setting, the essential features of a systematic-aware density estimation problem: a nontrivial nominal density and interpretable nuisance deformations.

\subsection{Toy dataset}
Each event is described by a binary class label $c \in \{A, B\}$ (drawn with equal probability), a two-dimensional vector $x=(x_1,x_2)$, and a two-dimensional feature vector $y=(y_1,y_2)$, also called the ``score''. The pair $(c, x)$ provides the conditioning information, while $y$ is the observable whose density we model. The kinematics are drawn from two class-dependent Gaussian clusters,
\begin{equation}
x \mid c \sim \mathcal{N}\!\big(\mu_c,\ \mathrm{diag}(\sigma_c^2)\big), \qquad
\begin{aligned}
\mu_A &= (-0.5,\, 0), & \sigma_A &= (0.9,\, 0.6),\\
\mu_B &= (+0.5,\, 0), & \sigma_B &= (0.6,\, 0.4).
\end{aligned}
\end{equation}
The nominal feature density $p_{\text{nom}}(y \mid x, c)$ is a bivariate Gaussian whose mean, per-axis spread and correlation depend nontrivially on the kinematics,
\begin{align}
\mu(x) &= \big(\,\sin(1.5\,x_1) + 0.3\,x_2,\ \ 0.3\,x_1^2 - 1.2 + 0.5\sin(x_2)\,\big),\\
\sigma_1(x) &= \mathrm{softplus}(0.4\,x_1 + 0.1), \qquad \sigma_2(x) = \mathrm{softplus}(-0.2\,x_1 + 0.4),\\
\rho(x) &= 0.8\,\tanh\!\big(0.5\,(x_1 + x_2)\big).
\end{align}

\begin{figure}[htb]
  \centering
  \includegraphics[width=\textwidth]{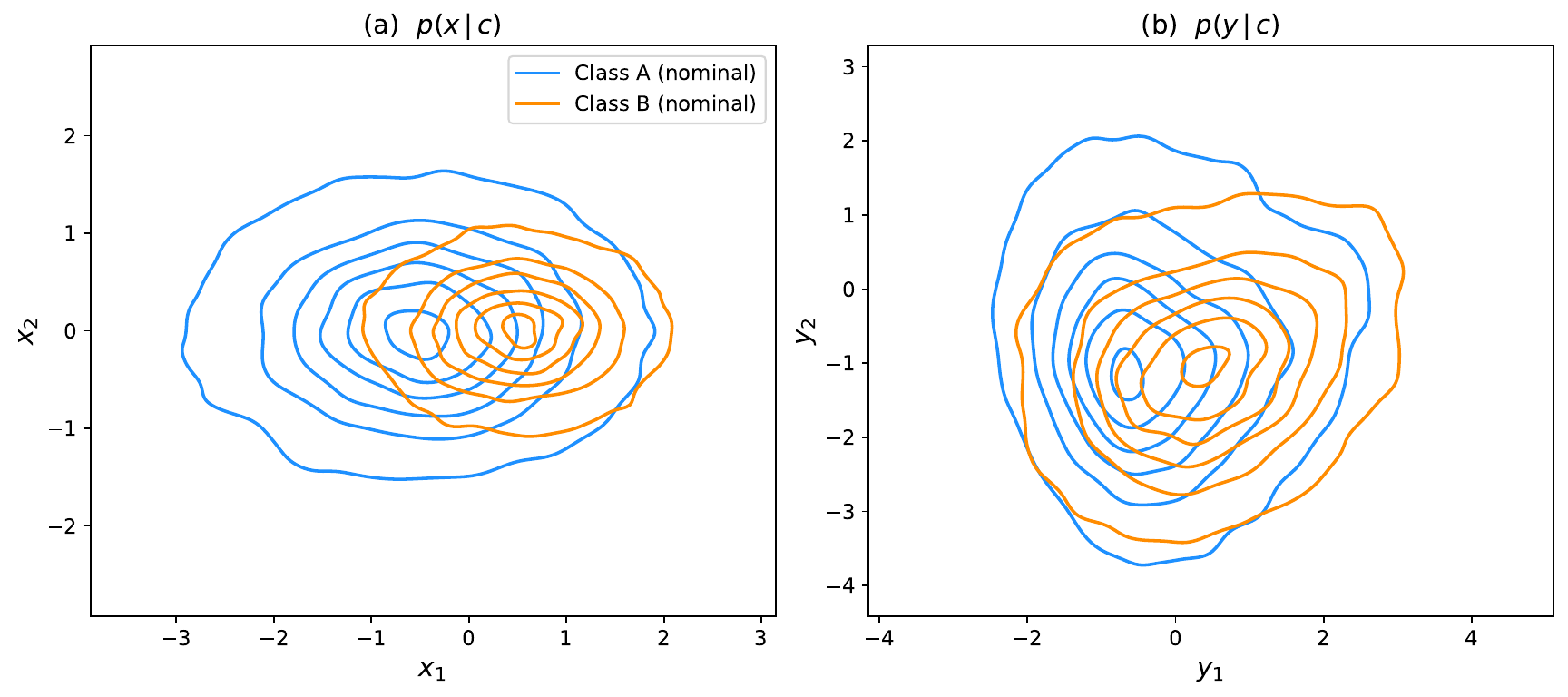}
  \caption{Nominal dataset at $\nu = 0$: the kinematic density $p(x)$ and the marginal score density $p(y \mid c)$, for class A (blue) and class B (orange). Both vary nontrivially between the two classes, illustrating the structure that the nominal flows must capture.}
  \label{fig:dataset}
\end{figure}

Figure~\ref{fig:dataset} illustrates the $p(x)$ density and the marginal $p(y \mid c)$.

\begin{figure}[tbp]
  \centering
  \includegraphics[width=\textwidth]{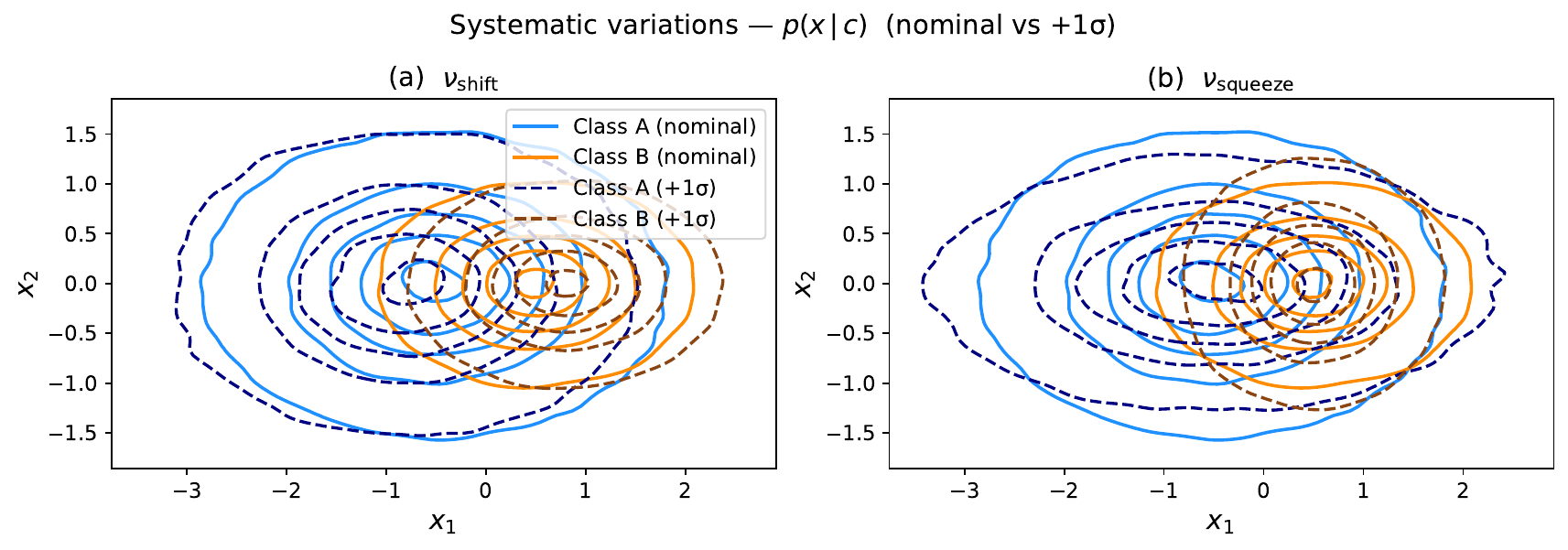}\\[1.5ex]
  \includegraphics[width=0.9\textwidth]{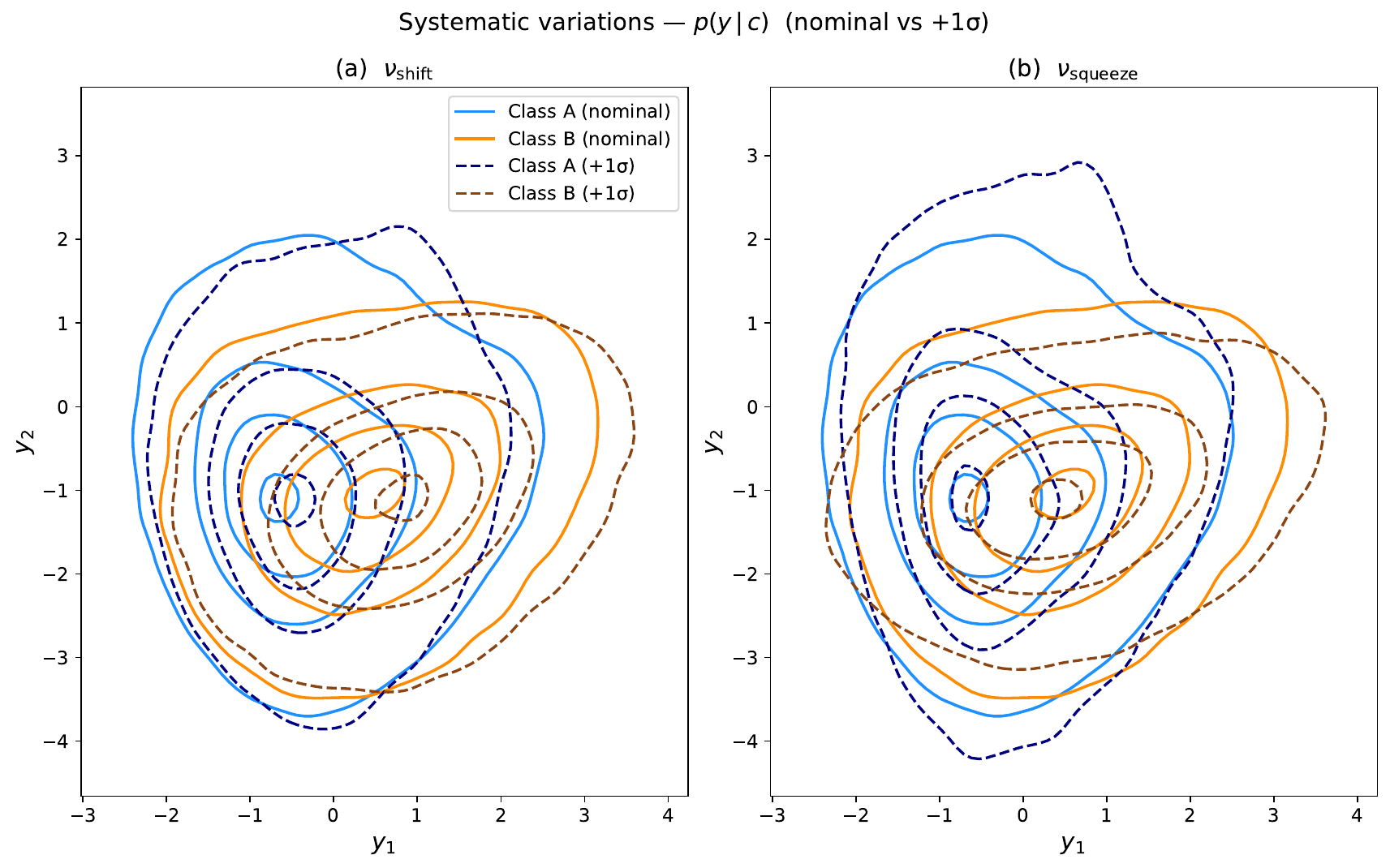}
  \caption{Effect of the two nuisance parameters, comparing the nominal distribution ($\nu = 0$, solid) with the $+1\sigma$ variation (dashed), for class A (blue) and class B (orange). \emph{Top:} effect on the kinematic density $p(x \mid c)$; $\nu_{\text{shift}}$ (left) translates the cluster centroids along $x_1$ with opposite sign for the two classes, while $\nu_{\text{squeeze}}$ (right) applies a volume-preserving, axis-anti-correlated squeeze of the kinematics about each centroid, with the sign mirrored between the two classes. \emph{Bottom:} effect on the marginal score density $p(y \mid c)$ (marginalized over the kinematics); $\nu_{\text{shift}}$ (left) adds a linear, $x$-dependent shift to the score mean, while $\nu_{\text{squeeze}}$ (right) rescales the score spread exponentially. In both cases the response has opposite sign for the two classes.}
  \label{fig:nuis_systematics}
\end{figure}

Two nuisance parameters drive the systematic variations, each acting with opposite sign on the two classes (anti-correlated response) and simultaneously on the kinematics and on the conditional feature density:
\begin{itemize}
  \item $\nu_{\text{shift}}$ translates the cluster centroids along $x_1$, $x \to x \mp \nu_{\text{shift}}\, s_{\text{shift}}\, \hat{d}$ (with $\hat{d}=(1,0)$; $-$ for class $A$ and $+$ for class $B$), and adds a linear, $x$-dependent shift to the score mean;
  \item $\nu_{\text{squeeze}}$ applies a volume-preserving ($\det = 1$) axis-anti-correlated squeeze of the kinematics about each centroid, $x \to \mu_c + \mathrm{diag}(e^{+\alpha}, e^{-\alpha})\,(x - \mu_c)$ with $\alpha = \nu_{\text{squeeze}}\, s_{\text{squeeze}}$ (the sign mirrored between classes), together with an exponential rescaling of the score spread.
\end{itemize}
Because both nuisances perturb the conditioning kinematics \emph{and} the score itself, they induce a genuine, interpretable dependence $p(y \mid x, c, \nu)$ that the FNF must capture. Figure~\ref{fig:nuis_systematics} shows their effect on the kinematic density $p(x \mid c)$ (top) and on the marginal score density $p(y \mid c)$ (bottom), comparing the nominal distribution with the $+1\sigma$ variation of the two nuisances.

The generator deformations are set by the kinematic scales $s_{\text{shift}}=0.3$, $s_{\text{squeeze}}=0.2$ and the feature-space response scales $0.3$ and $0.2$, so that both nuisances induce a sizeable, $x$-dependent deformation of the density. All quantitative results below are evaluated on a dataset of $10^5$ samples.

\subsection{Model architecture and training}
\label{sec:architecture}

Since both the kinematics $x$ and the score $y$ are affected by the nuisances, we factorize the joint density as $p(x, y \mid c, \nu) = p(x \mid c, \nu)\, p(y \mid x, c, \nu)$ and model each factor with a dedicated FNF. This split is a modeling choice rather than a necessity: a single FNF could model the full four-dimensional joint density $p(x, y \mid c, \nu)$ directly. We adopt it because it mirrors the common structure of a HEP analysis, where the conditioning kinematics $x$ are kept separate from the discriminating score $y$, so that each component can be modeled, trained, and validated on its own. Each factor is in turn the composition of a fixed \emph{nominal base flow}, trained once on the nominal ($\nu = 0$) simulation, and a learnable \emph{factorizable residual transformation} $T_\nu$ that pulls the observed sample back to the nominal reference, exactly as in Section~\ref{sec:factorizable_flows}.

\paragraph{Nominal base flows.} The nominal densities $p_{\text{nom}}(x \mid c)$ and $p_{\text{nom}}(y \mid x, c)$ are modeled with autoregressive Neural Spline Flows (NSF)~\cite{durkan2019neuralsplineflows}, using monotonic rational-quadratic spline transforms with $20$ bins. The kinematic flow uses $2$ transforms and is conditioned on the one-hot class label $c$; the score flow uses $3$ transforms and is conditioned on $(c, x)$. Both use masked MLPs with hidden layers $(128, 128, 128)$. They are trained by maximum likelihood (minimizing the negative log-likelihood) on nominal samples drawn on the fly from the generator at $\nu = 0$, and are then frozen.

\paragraph{Factorizable residual.} The residual transformation $T_\nu$ is implemented as described in Section~\ref{sec:factorizable_flows}. Each internal network's output layer is initialized to zero so that $T_\nu$ starts as the identity. For this experiment we use two independent nuisances and disable the pairwise cross-terms. We stack $2$ residual layers for the score flow, interleaved with fixed reverse permutations to mix the two feature dimensions, and $1$ layer for the kinematic flow; the per-nuisance masked MLPs use hidden layers $(128,128,128)$ for the score and $(128,128)$ for the kinematics. The architecture and hyperparameters are summarized in Table~\ref{tab:hyperparams}. All models are implemented in PyTorch~\cite{Paszke:2019pytorch} on top of the \texttt{zuko} normalizing-flow library~\cite{rozet2022zuko}, and our implementation, together with the toy dataset, is publicly available~\cite{Valsecchi:2026fnf}.

\paragraph{Training the residual.} With the base flows frozen, only the residual layers are trained, again by maximum likelihood. Crucially, the residual is trained \emph{only} on the per-nuisance $\pm 1\sigma$ templates: for each nuisance we generate samples with that nuisance set to $\nu_k = \pm 1$ and all others at zero, yielding four template generators for the two nuisances ($\nu_{\text{shift}} = \pm 1$, $\nu_{\text{squeeze}} = \pm 1$). Each training batch mixes equal-size chunks from the four templates, labeled by their nuisance vector $\nu$, and the model maximizes their likelihood under $p(\cdot \mid x, c, \nu) = p_{\text{nom}} \circ T_\nu$. Because each nuisance is only ever seen in isolation, the factorized additive structure is what allows the joint, multi-nuisance response to be recovered at inference by summation, never requiring training samples from the combined $K$-dimensional variation space. We assess the validity of this factorization assumption, and the effect of enabling the pairwise cross-terms, in Section~\ref{sec:additivity}.

\paragraph{Optimization.} All flows are optimized with AdamW~\cite{AdamW} at a learning rate of $10^{-4}$ and batch size $1024$, using a linear warm-up over the first epoch followed by cosine decay. The nominal base flows are trained for $100$ epochs and the residual transformations for $60$ epochs, each epoch comprising $1000$ steps.

\begin{table}[!htbp]
  \centering
  \begin{tabular}{lccccc}
    \toprule
    Component & Type & Features & Context & Layers & Hidden \\
    \midrule
    Nominal $p_{\text{nom}}(x \mid c)$     & NSF ($20$ bins)   & $2$ & $2$ & $2$ & $(128,128,128)$ \\
    Nominal $p_{\text{nom}}(y \mid x, c)$  & NSF ($20$ bins)   & $2$ & $4$ & $3$ & $(128,128,128)$ \\
    FNF residual (kinematic)               & affine AR         & $2$ & $2$ & $1$ & $(128,128)$ \\
    FNF residual (score)                   & affine AR         & $2$ & $4$ & $2$ & $(128,128,128)$ \\
    \bottomrule
  \end{tabular}
  \caption{Architecture and training hyperparameters of the FNF model for the toy experiment. ``Layers'' counts spline transforms for the base flows and residual layers for the FNF residuals. The base flows are conditioned on the context only; the residuals additionally take the $2$ nuisances $\nu$ as polynomial inputs.}
  \label{tab:hyperparams}
\end{table}

\subsection{Results}
\begin{figure}[tbp]
  \centering
  \begin{subfigure}{\textwidth}
    \centering
    \includegraphics[width=\textwidth]{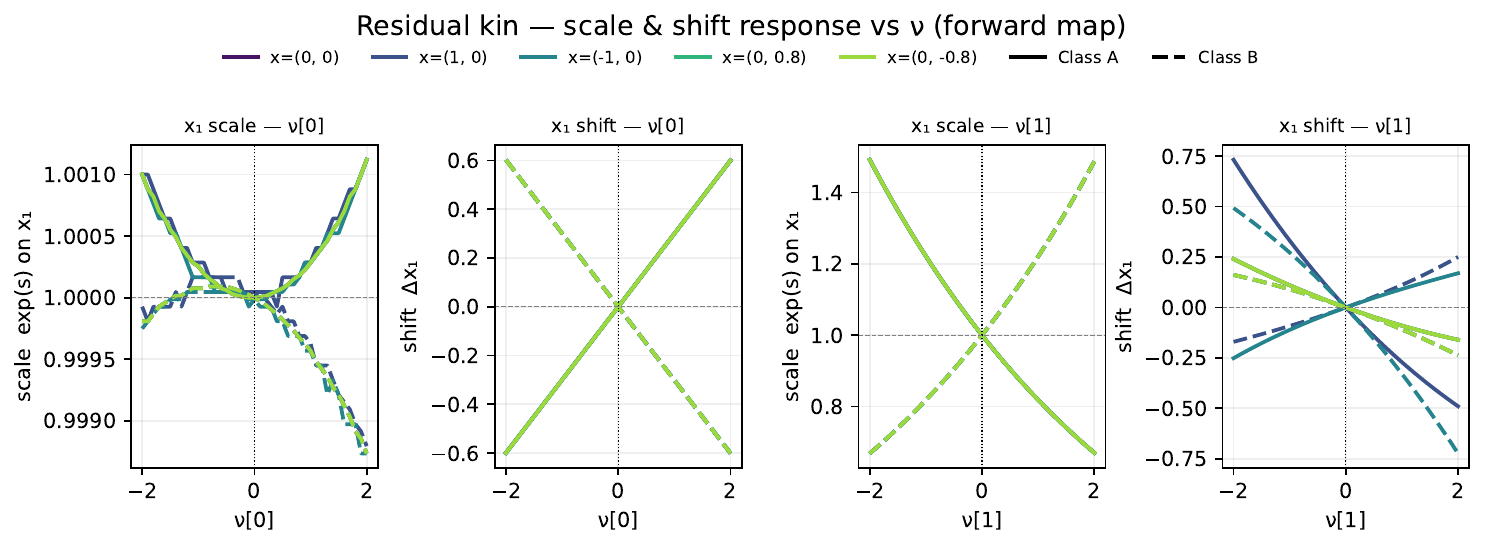}
    \caption{Learned affine response}\label{fig:residual_kin}
  \end{subfigure}
  \\[1ex]
  \begin{subfigure}{\textwidth}
    \centering
    \includegraphics[width=\textwidth]{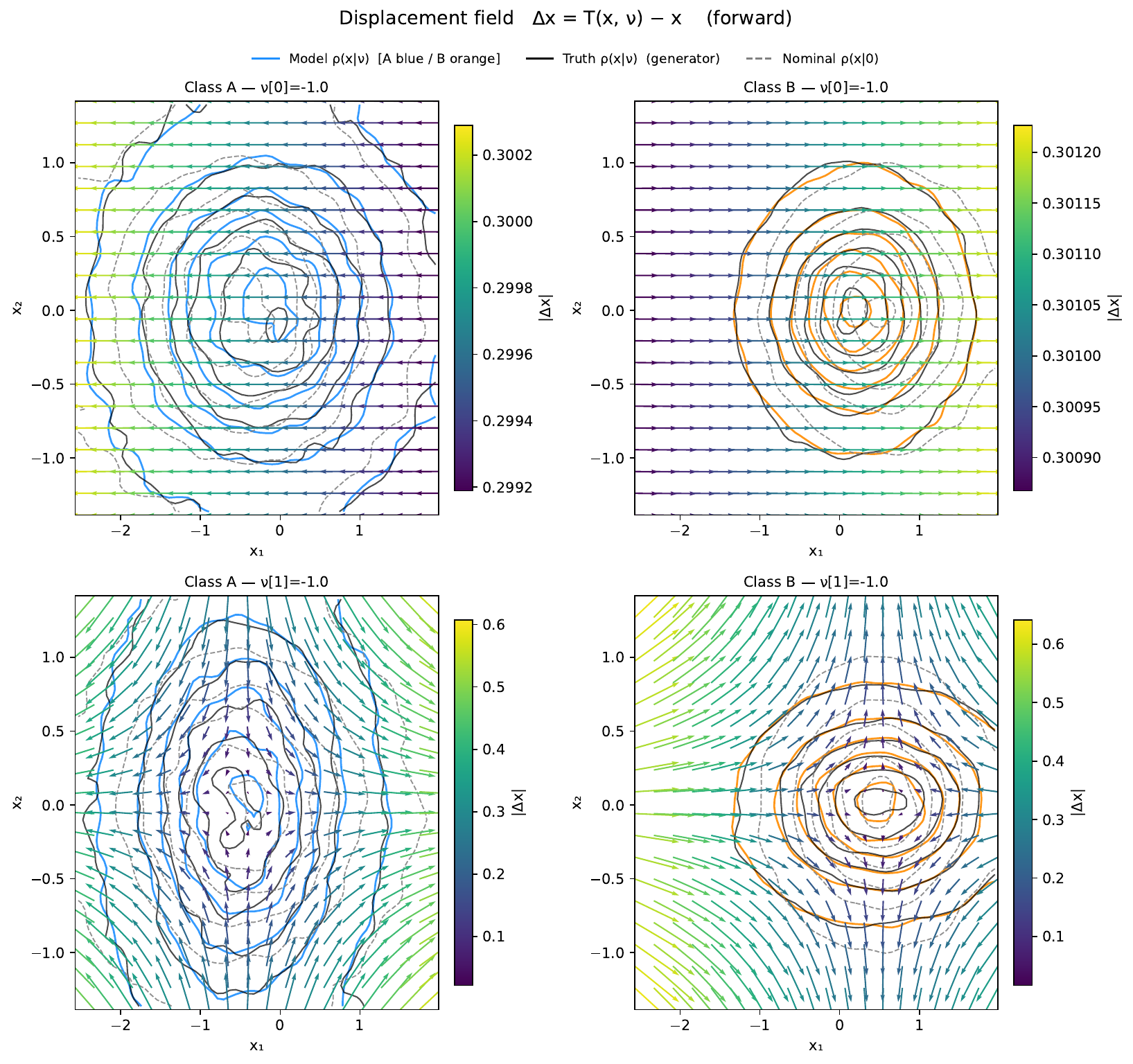}
    \caption{Displacement field $\Delta x$}\label{fig:residual_kin_transfer}
  \end{subfigure}
  \caption{Validation of the factorizable residual for the kinematic density $p(x \mid c, \nu)$. \textbf{(a)} Learned affine response of the first component $x_1$: the scale factor $\exp(s)$ and the shift $t$ versus $\nu_{\text{shift}}$ (left pair) and $\nu_{\text{squeeze}}$ (right pair), at several representative points $x$ and for both classes. For $\nu_{\text{shift}}$ the scale stays at unity ($s \approx 0$) and the shift is linear, as expected for a pure translation; for $\nu_{\text{squeeze}}$ both are nontrivial and $x$-dependent. The component $x_2$ responds analogously for $\nu_{\text{shift}}$ and with opposite sign for $\nu_{\text{squeeze}}$, reflecting the axis-anti-correlated squeeze. \textbf{(b)} Displacement field $\Delta x = T_\nu(x \mid c, \nu) - x$ at $\nu = -1$ for the $\nu_{\text{shift}}$ (top) and $\nu_{\text{squeeze}}$ (bottom) deformations, class A (left) and class B (right); arrows colored by $|\Delta x|$. Solid contours show the deformed model density $p(x \mid \nu)$ (blue for A, orange for B) over the generator truth (black), dashed gray the nominal density $p(x \mid 0)$. The residual transports the deformed distribution back onto the nominal reference.}
  \label{fig:residual_kin_combined}
\end{figure}

We first assess the FNF applied to the kinematic density $p(x \mid c, \nu)$. Figure~\ref{fig:residual_kin} shows the learned affine response of the residual transformation, the scale factor $\exp(s)$ and the shift $t$, evaluated at several representative points in $x$ for both nuisances. For the $\nu_{\text{shift}}$ deformation the scale remains at unity (i.e.\ the log-scale $s \approx 0$), while the shift grows linearly with $\nu_{\text{shift}}$, exactly as expected for a pure translation. For the $\nu_{\text{squeeze}}$ deformation, instead, both the scale and the shift are nontrivial and depend on the specific point in $x$, as expected for the volume-preserving squeeze. Figure~\ref{fig:residual_kin_transfer} shows the corresponding displacement field $\Delta x = T_\nu(x \mid c, \nu) - x$ encoded by the residual: the transformation maps the deformed distribution back onto the nominal reference (dashed gray contours), demonstrating that the model learns a smooth deformation that recovers the nominal density.

\begin{figure}[tbp]
  \centering
  \begin{subfigure}{\textwidth}
    \centering
    \includegraphics[width=\textwidth]{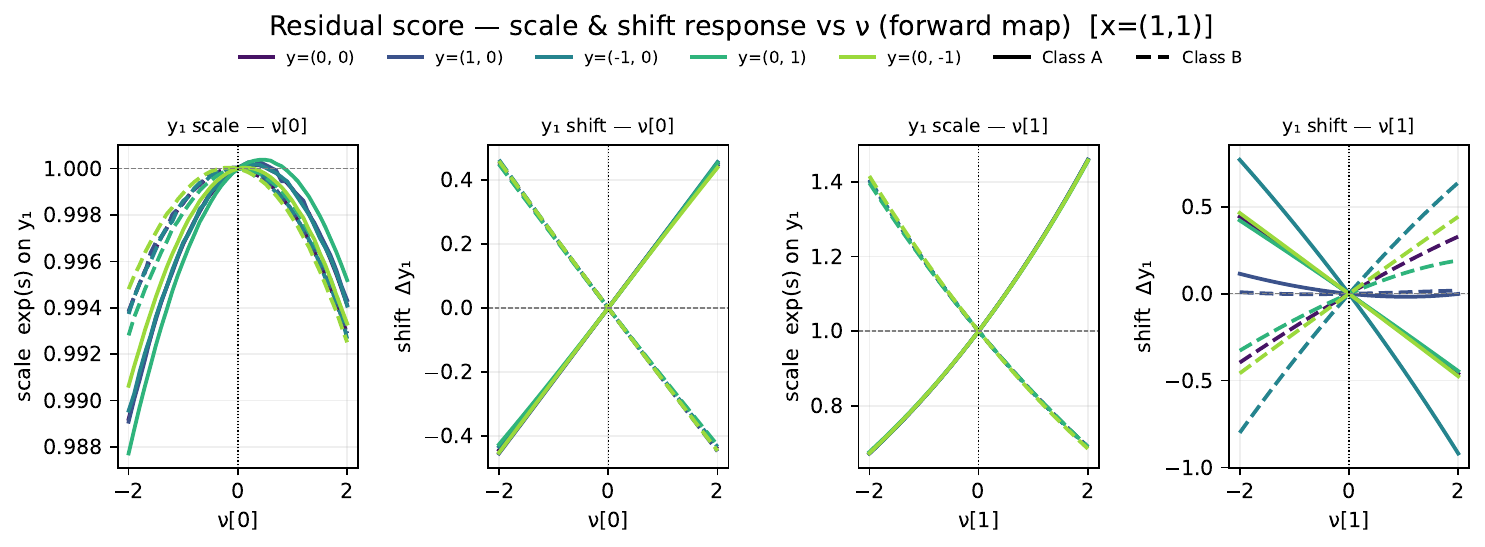}
    \caption{Learned affine response}\label{fig:residual_score}
  \end{subfigure}
  \\[1ex]
  \begin{subfigure}{\textwidth}
    \centering
    \includegraphics[width=\textwidth]{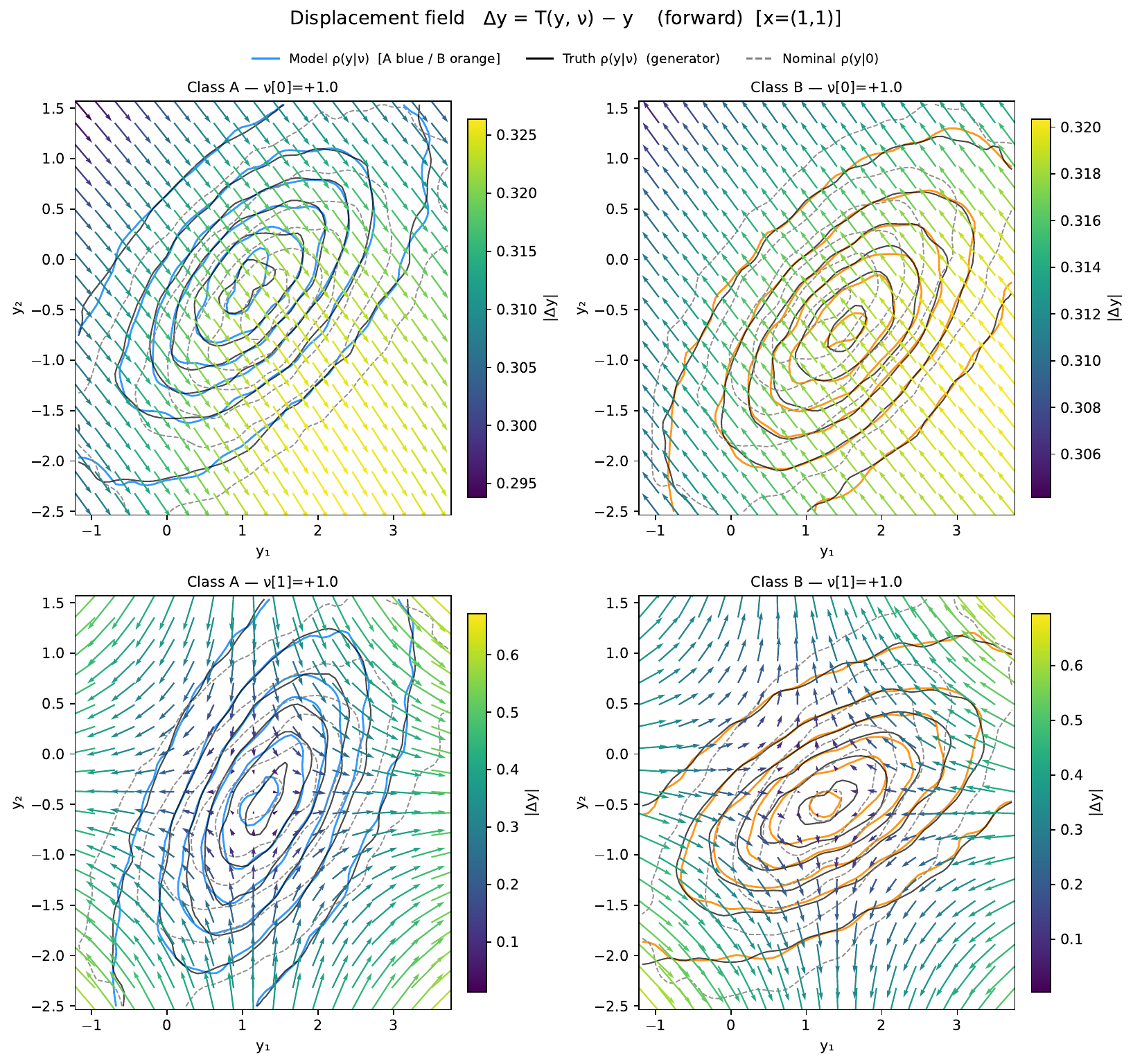}
    \caption{Displacement field $\Delta y$}\label{fig:residual_score_transfer}
  \end{subfigure}
  \caption{Validation of the score residual for $p(y \mid x, c, \nu)$ at the conditioning point $x = (1,1)$. \textbf{(a)} Learned affine response of the first component $y_1$: the scale factor $\exp(s)$ and the shift $t$ versus $\nu_{\text{shift}}$ (left pair) and $\nu_{\text{squeeze}}$ (right pair), at four representative points in $y$ and for both classes; the component $y_2$ behaves analogously and is omitted. \textbf{(b)} Induced displacement field $\Delta y = T_\nu(y \mid x, \nu) - y$ for $+1\sigma$ variations of the $\nu_{\text{shift}}$ (top) and $\nu_{\text{squeeze}}$ (bottom) nuisances, class A (left) and class B (right); arrows colored by $|\Delta y|$. Solid contours show the deformed model density (blue for A, orange for B) and the generator truth (black), dashed gray the nominal density $p(y \mid 0)$. The residual maps the distorted score distribution back onto the nominal reference for both classes.}
  \label{fig:residual_score_combined}
\end{figure}

We then validate the residual on the conditional score density $p(y \mid x, c, \nu)$. This is the more stringent test: unlike the kinematics, the score is correlated and its nominal shape varies with the conditioning kinematics, so the residual must reproduce an $x$-dependent deformation. Figure~\ref{fig:residual_score} shows the learned affine response of $y_1$ at the conditioning point $x=(1,1)$, evaluated at four representative points in $y$ and for both classes.

The two nuisances leave qualitatively distinct signatures, and the residual assigns each to the correct affine component. Under $\nu_{\text{shift}}$ the scale stays close to unity while the shift grows linearly, reproducing the pure translation of the score mean that defines this nuisance. Under $\nu_{\text{squeeze}}$ the response is instead carried by the scale, whose logarithm grows linearly with the nuisance (an exponential rescaling of the score spread), accompanied by a smaller, $y$-dependent shift. The two classes respond with opposite sign throughout (solid versus dashed), recovering the anti-correlated class dependence built into the generator, while the spread of the curves across the four $y$ points shows that the learned response is local rather than a single global affine, as the autoregressive networks permit.

Figure~\ref{fig:residual_score_transfer} confirms this at the level of the full distribution. At $+1\sigma$ the displacement field $\Delta y = T_\nu(y \mid x, \nu) - y$ is a near-uniform translation for $\nu_{\text{shift}}$ and a convergent or divergent squeeze for $\nu_{\text{squeeze}}$, with opposite orientation for the two classes. In every case the deformed model density (solid) overlays the generator truth (black), and the residual transports it back onto the nominal reference (dashed gray), demonstrating closure not only in the learned affine parameters but in the morphed score distribution itself.

\subsection{Additivity and correlated effects}
\label{sec:additivity}
The results above validate each nuisance in isolation. We now turn to the combined response, with both nuisances active simultaneously. By construction, the factorized FNF reconstructs the joint deformation as the sum of the per-nuisance scale and shift contributions, so that the scale factors compose multiplicatively across nuisances. Each nuisance's own response is retained up to second order through its linear and quadratic terms; the additive structure across nuisances, with the cross-terms omitted, encodes the assumption that distinct systematic sources act independently on the density. It is exact whenever that assumption holds and deviates only through the neglected pairwise interactions. This is precisely the factorization underlying the standard template-variation approach, where each uncertainty is described by its own independent variations and correlated shape effects between sources are not modeled at the template level.

To quantify the impact of this approximation, we compare two configurations: (i) the \emph{factorized} model, with cross-terms disabled and each nuisance trained on its own $\pm 1\sigma$ samples; and (ii) the \emph{correlated} model, with the pairwise cross-terms $\phi$ and $\psi$ enabled and trained on additional samples in which both nuisances are varied simultaneously. For both we quantify the goodness of fit across the full nuisance plane through the per-event excess negative log-likelihood with respect to the optimal (truth) likelihood. Averaged over events generated by the truth at a given $\nu$, this excess is the Kullback-Leibler (KL) divergence
\begin{equation}\label{eq:kl}
\Delta\mathrm{NLL}(\nu) \equiv \mathrm{KL}\!\left[\,p_{\text{truth}}\,\big\|\,p_{\text{model}}\,\right]
= \mathbb{E}_{p_{\text{truth}}}\!\left[\,\log p_{\text{truth}}(x, y \mid c, \nu) - \log p_{\text{model}}(x, y \mid c, \nu)\,\right] \ge 0,
\end{equation}
where the expectation runs over events $(c, x, y)$ drawn from the truth at $\nu$. It vanishes if and only if the model matches the truth, giving an absolute, per-event measure of fit quality in nats.

\begin{figure}[t]
  \centering
  \includegraphics[width=0.49\textwidth]{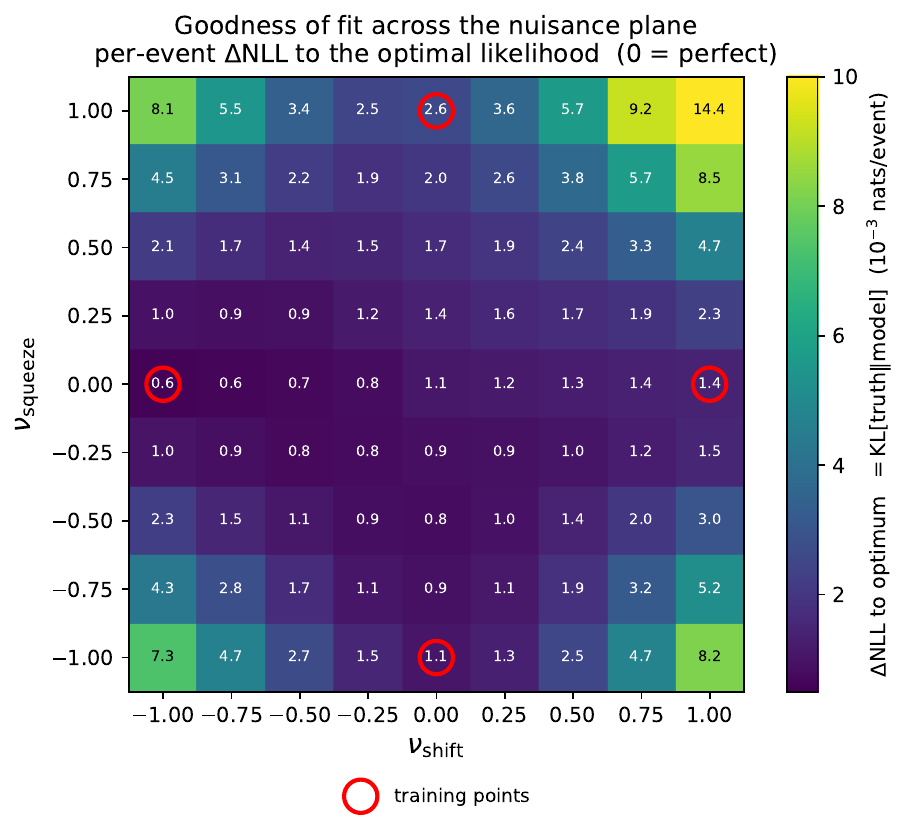}
  \hfill
  \includegraphics[width=0.49\textwidth]{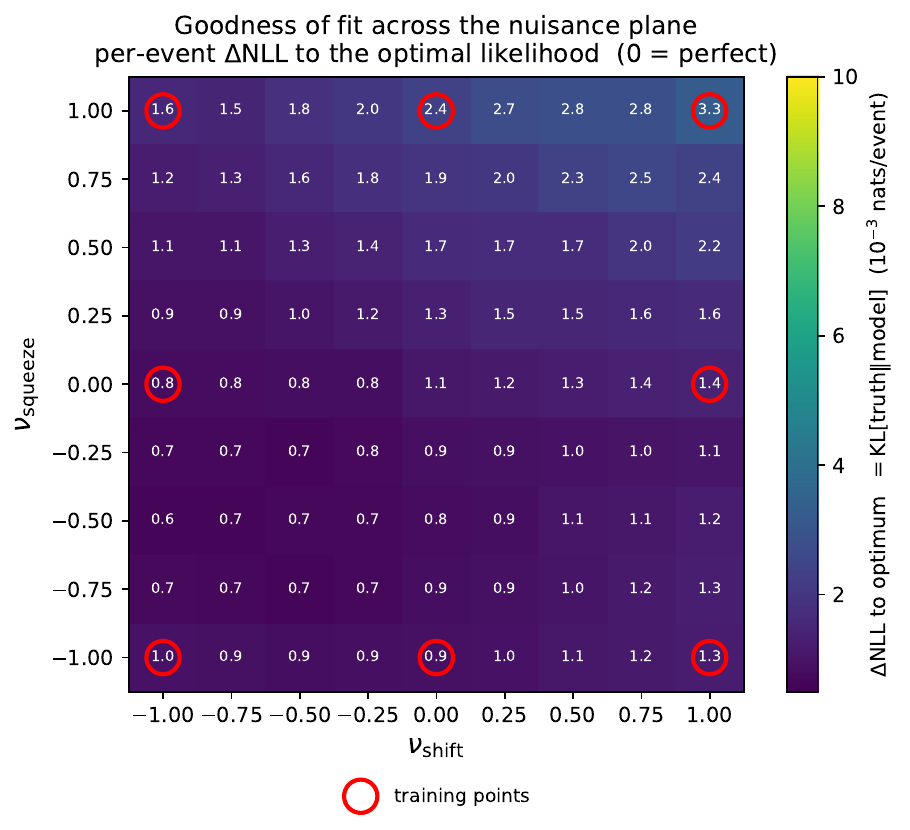}
  \caption{Goodness of fit across the nuisance plane, measured by the per-event $\Delta\mathrm{NLL} = \mathrm{KL}[\,p_{\text{truth}}\,\|\,p_{\text{model}}\,]$ (Eq.~\eqref{eq:kl}; in $10^{-3}$ nats per event, $0$ being the optimum). Red circles mark the nuisance configurations used for training. \emph{Left:} factorized model (cross-terms disabled), trained only on the four single-nuisance $\pm1\sigma$ variations. \emph{Right:} correlated model (cross-terms enabled), additionally trained on the four combined variations at the corners of the plane.}
  \label{fig:gof_nuisance_plane}
\end{figure}

Figure~\ref{fig:gof_nuisance_plane} compares the two configurations across the nuisance plane. The factorized model reproduces the optimal likelihood to within a few times $10^{-3}$ nats per event near the training axes and around the nominal point. Away from the axes, however, the missing interaction grows steadily, exceeding $10^{-2}$ in the corners where both nuisances are large and act simultaneously. Enabling the cross-terms, constrained by the additional combined-variation samples, suppresses this residual across the whole plane, which stays below $3.3\times10^{-3}$ nats per event, an improvement of more than a factor of four in the most correlated region. The factorized approximation is thus adequate when the nuisances are moderate or well constrained, whereas the cross-terms become necessary to preserve fidelity under large, simultaneous variations.

\begin{figure}[tbp]
  \centering
  \includegraphics[width=\textwidth]{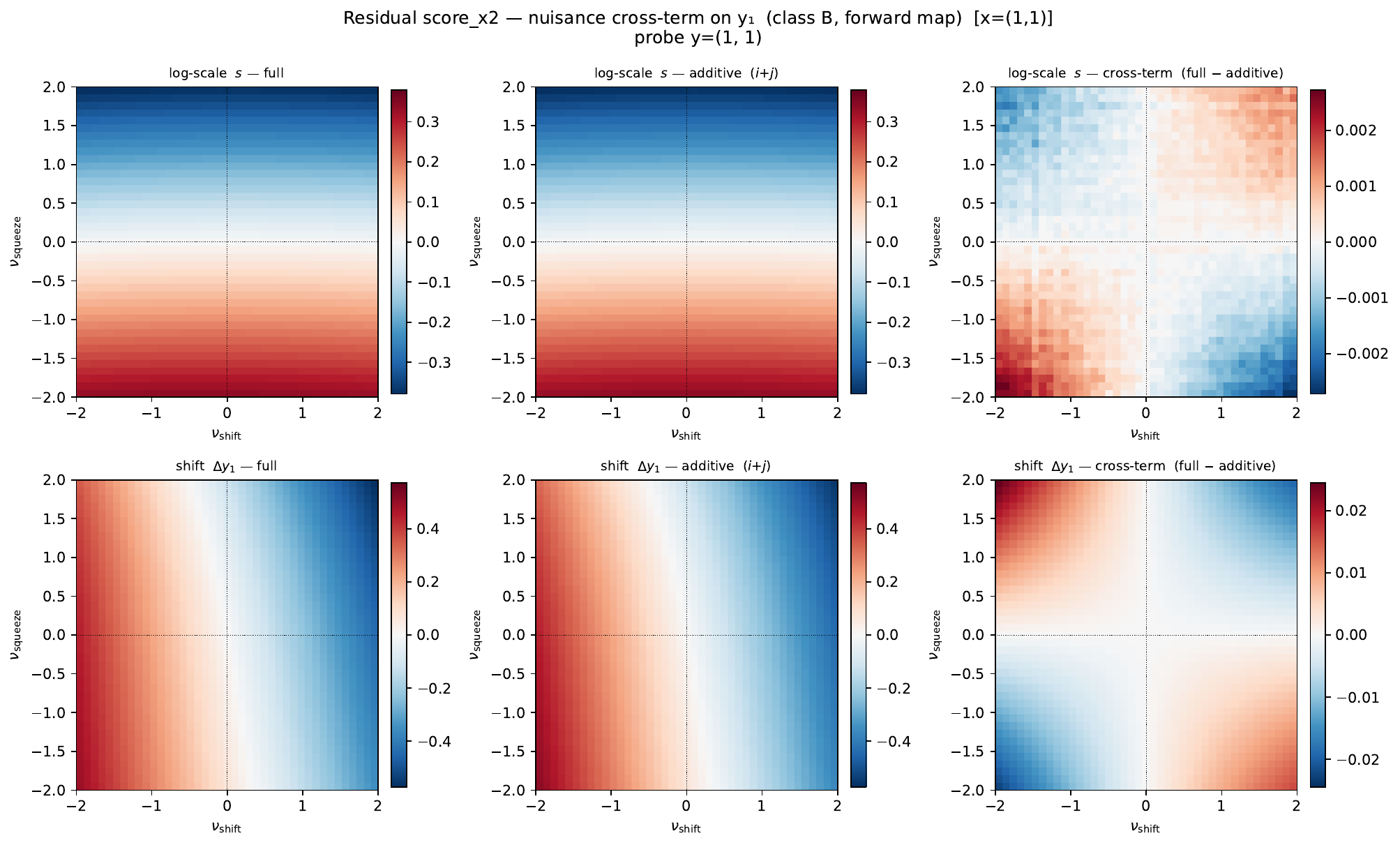}
  \caption{Direct view of the learned cross-term for the score residual ($y_1$, class B, conditioning $x=(1,1)$, probe point $y=(1,1)$). Columns show, for the log-scale $s$ (top) and the shift $\Delta y_1$ (bottom), the full correlated response (left), its additive part, i.e.\ the sum of the per-nuisance contributions (middle), and the residual cross-term, full minus additive (right). The cross-term carries the bilinear $\nu_{\text{shift}}\,\nu_{\text{squeeze}}$ structure, vanishing on the axes, changing sign across quadrants, and growing toward the corners, while remaining one to two orders of magnitude below the full response.}
  \label{fig:crossterm}
\end{figure}

Figure~\ref{fig:crossterm} makes this interaction explicit. Decomposing the correlated model's affine response into its additive part and the bilinear remainder, the cross-term is the small correction that the factorized model omits: it vanishes on the axes, where a single nuisance is active, follows the $\nu_{\text{shift}}\,\nu_{\text{squeeze}}$ product across the plane, and is largest in the corners. Its magnitude stays one to two orders of magnitude below the full response, which is why the factorized approximation holds for moderate or single-nuisance variations and breaks down only where both nuisances are simultaneously large, exactly the region where the goodness of fit in Figure~\ref{fig:gof_nuisance_plane} degrades.

These two configurations span the available precision-versus-cost trade-off. The factorized model is the cheapest: it requires only the per-nuisance variations and scales linearly with the number of uncertainties, exactly as in the established template approach. When residual correlations between sources are non-negligible, the FNF can account for them natively by enabling the cross-terms, at the cost of generating combined-variation samples and a number of interaction terms that grows with the number of nuisance pairs. The same architecture therefore interpolates between the standard factorized approximation and a fully correlated model, with the level of approximation chosen explicitly.

\section{Discussion}
\label{sec:discussion}
This formulation has several key features:
\begin{itemize}
    \item \textbf{Factorized, modular structure}: Each nuisance $\nu_k$ is handled by a separate network $\Psi^{k}$, so its effect is modeled independently (when excluding cross-terms). This lets us train each variation from its own $\pm 1\sigma$ samples alone, avoids the combinatorial explosion of sampling the full $K$-dimensional space, and allows new uncertainties to be added without retraining the existing model.

    \item \textbf{Interpretability}: The learned coefficients carry a direct meaning: $\alpha^{k}$ and $\gamma^{k}$ give the linear response of the feature distribution to $\nu_k$, while $\beta^{k}$ and $\delta^{k}$ capture the quadratic effects.

    \item \textbf{Efficiency}: The joint transformation is evaluated in a single forward pass by summing the per-nuisance outputs, and its triangular Jacobian makes the density fast to evaluate during the fit.

    \item \textbf{Expressivity}: Despite the factorized structure, the transformation remains highly expressive: the neural-network coefficients and the autoregressive structure capture complex, non-linear dependencies on the features $y$ and kinematics $x$.
\end{itemize}
  
\paragraph{Interplay with Optimal Transport}
The morphing transformation encoded by the Factorizable Normalizing Flow is not guaranteed to be ``optimal'' in the Optimal Transport (OT) sense, i.e., the unique map that transports the probability mass between the two distributions at minimal cost~\cite{Peyre:2019cot}. This is not a limitation of our approach. On the contrary, by letting the solution live in a larger space, without imposing uniqueness or minimal-cost constraints, the FNF can be trained directly by maximum likelihood, whereas constructing an explicit OT map, for example through convex potentials or continuous flows, is typically far more computationally expensive~\cite{Onken:2021otflow,huang2021convexpotentialflowsuniversal,amos2017inputconvexneuralnetworks,Tong:2024otcfm}. Optimal transport has nonetheless found fruitful application in high energy physics, for instance to define a metric on the space of collider events~\cite{Komiske:2019fks} and to map between distributions for calibration and inference~\cite{ATLAS:2025calib,Algren:2025gap}.

\section{Related work}
\label{sec:related_work}
Encoding systematic uncertainties in unbinned likelihood fits is a crucial topic that has been addressed from several directions. A broad family of methods falls under simulation-based inference~\cite{Cranmer:2020wdu,Bahl:2024meb}, where the intractable likelihood, or the likelihood ratio, is learned directly from simulated samples rather than modeled analytically. In the effective field theory (EFT) context, the known parametric dependence of the matrix element can be exploited to learn per-event likelihood ratios and scores~\cite{Brehmer:2018eca}. Closely related to our approach are methods that parameterize the density or the likelihood ratio with generative models trained on simulation: Normalizing Flows have been used to unify simulation and inference~\cite{Krause:2025unifying}, generative models enable data-driven high-dimensional inference~\cite{Amram:2025hisigma}, and machine-learned systematic uncertainties have been incorporated into unbinned cross-section measurements~\cite{zwzt-1rrw} and global SMEFT analyses~\cite{GomezAmbrosio:2022step}.

The ATLAS Collaboration has recently deployed a neural SBI technique for parameter estimation~\cite{ATLAS:2025nsbi}, in which neural-network classifiers are trained to discriminate between samples generated under different parameter hypotheses. The resulting parameterized classifier yields the per-event likelihood ratio as a continuous function of the parameters of interest and the nuisance parameters, enabling fully unbinned inference on high-dimensional observables; the technique was applied to the measurement of off-shell Higgs boson production in the $H \to ZZ \to 4\ell$ channel~\cite{ATLAS:2025higgs}. A complementary, tree-based strategy is the refinable modeling of Ref.~\cite{Schofbeck:2025smeft}, where boosted information trees learn the parametric dependence of the likelihood ratio as a refinement of a base model for unbinned SMEFT analyses, replacing the neural-network estimator with a gradient-boosted tree ensemble.

These approaches share the same underlying idea as our work: the effect of the parameters is encoded as a \emph{parametric deformation} of the model, learned from simulation and interpolated smoothly across the parameter space. The key distinction is \emph{what} is deformed and \emph{how} it is learned. Both the ATLAS and the tree-based methods parameterize the \emph{likelihood ratio} and learn it discriminatively with classifiers, neural networks in one case and boosted decision trees in the other. In contrast, the FNF parameterizes the \emph{density} itself through an invertible normalizing-flow transformation, providing an explicit and normalized model of $p(y \mid x, \nu)$ whose deformation is interpretable and whose Jacobian makes the likelihood directly tractable.  

\subsection{Future directions}
Factorizable Normalizing Flows are expected to find many direct applications in future unbinned likelihood-fit workflows. They are particularly well suited to the methodology of HEP analyses, but they are a general density-estimation tool that could also prove useful in other fields that already rely heavily on Normalizing Flows for inference, such as cosmology~\cite{Dai:2023lcb} and gravitational-wave astrophysics~\cite{Dax:2021tsq}. Within HEP, we foresee direct applications to detector calibration~\cite{Daumann:2024flow}, unbinned parameter estimation, and unfolding~\cite{Butter:2025dm,Butter:2025analysis,butter2025simulationpriorindependentneuralunfolding,T2K:2025omnifold,Andreassen_2020}. In particular, the use of the FNF likelihood to profile nuisance parameters and extract parameters of interest in a full unbinned fit is the subject of a dedicated forthcoming publication.

\section{Conclusions}
\label{sec:conclusions}
We have introduced the Factorizable Normalizing Flow (FNF), a generative framework for modeling how a probability density deforms under continuous parameters such as systematic uncertainties. By decoupling a fixed, high-fidelity nominal flow from a learnable transformation that is polynomial in the parameters and factorized over them, the FNF generalizes the classical template-variation approach to a continuous, high-dimensional, and fully differentiable setting. Its central practical advantage is that each variation is learned independently from only its own $\pm 1\sigma$ samples, avoiding the combinatorial cost of sampling the joint parameter space, while the joint response is recovered by summing the per-parameter contributions at inference.

On a controlled problem with two interpretable nuisances acting jointly on the kinematics and on the conditional score, the learned scale and shift responses reproduce the imposed deformations and transport the distorted distributions back onto the nominal reference. Quantitatively, the factorized model attains closure to the optimal likelihood across the trained region, and enabling the optional cross-terms recovers the residual correlations where parameters vary strongly together, so that a single architecture spans the full range from the standard factorized approximation to a fully correlated model, with the level of approximation chosen explicitly. The resulting model is interpretable, scales linearly with the number of parameters, and is efficient to evaluate, with a tractable likelihood suited to profiling. By making systematic-aware density estimation practical at scale, the FNF lowers a key barrier to the adoption of unbinned likelihood fits in high energy physics, and offers a general, reusable tool for any inference workflow in which probability densities must be morphed as a function of continuous parameters. Its deployment in a profiled, unbinned likelihood fit to extract parameters of interest is the subject of dedicated forthcoming work.

\funding{This work was supported by the Swiss National Science Foundation under contract number 10003769.}

\roles{\textbf{D. Valsecchi}: Conceptualization, Methodology, Software, Formal analysis, Visualization, Writing (original draft). \textbf{M. Doneg\`a}: Conceptualization, Investigation, Writing (review and editing). \textbf{R. Wallny}: Supervision, Funding acquisition}

\data{The code implementing the Factorizable Normalizing Flow and the toy dataset shown in this paper is publicly available on Zenodo~\cite{Valsecchi:2026fnf}.}

\bibliographystyle{unsrtnat}
\bibliography{my_bibliography}

\end{document}

%% file: figs/fnf_schema_tikz.tex
\definecolor{nomPurple}{RGB}{106,90,205}%
\definecolor{defOrange}{RGB}{225,95,60}%
\definecolor{nuFill}{RGB}{253,235,200}%
\definecolor{nuBorder}{RGB}{214,158,70}%
\definecolor{nuText}{RGB}{176,108,22}%
\definecolor{featFill}{RGB}{231,231,231}%
\definecolor{featBorder}{RGB}{150,150,150}%
\definecolor{psiFill}{RGB}{224,220,245}%
\definecolor{psiBorder}{RGB}{138,127,208}%
\definecolor{abFill}{RGB}{214,236,224}%
\definecolor{abBorder}{RGB}{70,160,120}%
\definecolor{abText}{RGB}{27,118,88}%
\definecolor{gdFill}{RGB}{248,221,213}%
\definecolor{gdBorder}{RGB}{200,90,70}%
\definecolor{gdText}{RGB}{176,58,44}%
\definecolor{layerFill}{RGB}{246,243,238}%
\definecolor{layerBorder}{RGB}{182,176,166}%
\definecolor{bottomFill}{RGB}{244,247,244}%
\definecolor{bottomBorder}{RGB}{182,190,182}%
\definecolor{scalePanel}{RGB}{224,240,230}%
\definecolor{scaleBorder}{RGB}{120,180,150}%
\definecolor{shiftPanel}{RGB}{250,231,225}%
\definecolor{shiftBorder}{RGB}{220,140,120}%
\definecolor{warpPanel}{RGB}{236,234,248}%
\definecolor{warpBorder}{RGB}{160,150,210}%
\definecolor{axisGray}{RGB}{110,110,110}%

\begin{tikzpicture}[
    >={Stealth[length=2.5mm]},
    font=\small,
    block/.style={rounded corners=3pt, draw, thick, align=center,
                  inner sep=4pt, minimum height=8mm, minimum width=15mm},
    title/.style={font=\bfseries},
    capt/.style={font=\footnotesize, text=black!75},
  ]


\begin{scope}[shift={(0,0)}]
  \coordinate (cN) at (1.45,1.55);                 
  \draw[axisGray,->] (0,0) -- (0,3.1);             
  \draw[axisGray,->] (0,0) -- (3.1,0);             
  \foreach \r in {0.35,0.72,1.08,1.42}
     \draw[nomPurple,thick] (cN) circle (\r);
  \node[axisGray] at (-0.25,3.0) {$y_2$};
  \node[axisGray] at (3.0,-0.25) {$y_1$};
  \node[title] at (1.55,3.6) {Nominal ($\nu=0$)};
  \node[capt] at (1.55,-0.7) {$p_{\mathrm{nom}}(y)$ --- frozen from simulation};
\end{scope}

\draw[->,thick] (4.4,1.6) -- node[above,font=\footnotesize]{$y_{\mathrm{nom}}$} (3.4,1.6);

\begin{scope}[shift={(5.0,0)}]
  \node[block, fill=nuFill,   draw=nuBorder,   text=nuText]   (nu)  at (4.55,2.25) {$\nu$\\[-1pt]\footnotesize nuisance};
  \node[block, fill=featFill, draw=featBorder]                (y)   at (4.55,0.85) {$y$\\[-1pt]\footnotesize feature};
  \node[block, fill=psiFill,  draw=psiBorder]                 (psi) at (2.55,1.55) {$\Psi$\\[-1pt]\footnotesize masked MLP};
  \node[block, fill=abFill,   draw=abBorder,   text=abText]   (ab)  at (0.55,2.25) {$\alpha(y)$\\$\beta(y)$};
  \node[block, fill=gdFill,   draw=gdBorder,   text=gdText]   (gd)  at (0.55,0.85) {$\gamma(y)$\\$\delta(y)$};

  \draw[->] (y) -- (psi);
  \draw[->] (psi) -- (ab);
  \draw[->] (psi) -- (gd);
  \draw[nuBorder,dashed,->] (nu.west) -- (ab.east);

  \node[font=\footnotesize, align=left] at (2.3,-0.05)
        {$s=\alpha\nu+\beta\nu^2$\\$t=\gamma\nu+\delta\nu^2$};

  \begin{scope}[on background layer]
    \node[rounded corners=6pt, draw=layerBorder, fill=layerFill, thick,
          fit={(nu)(y)(psi)(ab)(gd)(2.3,-0.30)},
          inner sep=10pt] (layer) {};
  \end{scope}
  \node[title] at (2.55,3.6) {FNF layer};
\end{scope}

\draw[->,thick] (11.95,1.6) -- node[above,font=\footnotesize]{apply $T_\nu$} (10.85,1.6);

\begin{scope}[shift={(12.35,0)}]
  \def\cx{1.50}\def\cy{1.25}
  \def\sx{0.82}\def\sy{0.55}\def\sh{0.45}\def\kk{0.55}
  \draw[axisGray,->] (0,0) -- (0,3.1);
  \draw[axisGray,->] (0,0) -- (3.1,0);
  \foreach \r in {0.45,0.82,1.15,1.48}
     \draw[defOrange,thick,dashed,smooth]
        plot[domain=0:360,samples=96,variable=\t]
        ({\cx + \r*cos(\t)*\sx + \sh*(\r*sin(\t)*\sy)},
         {\cy + \r*sin(\t)*\sy + \kk*(\r*cos(\t)*\sx)^2});
  \node[axisGray] at (-0.25,3.0) {$y_2$};
  \node[axisGray] at (3.0,-0.25) {$y_1$};
  \node[title] at (1.55,3.6) {Deformed ($\nu\neq0$)};
  \node[capt] at (1.55,-0.7) {$p(y\mid\nu)$ --- locally distorted};
\end{scope}

\draw[axisGray!70, dotted, line width=1pt, -{Stealth[length=2.2mm]}]
     (2.5,3.95) to[bend left=10]
     node[midway, above=1pt, font=\footnotesize, text=axisGray]
          {systematic $\nu$ deforms the density}
     (13.4,3.95);

\node[font=\footnotesize, text=black!80, align=center] at (7.55,-1.1)
      {$T_\nu:\ y\ \longmapsto\ y_{\mathrm{nom}}=y\cdot e^{s}+t$\\[-1pt]
       \scriptsize\textcolor{axisGray}{pull each observed event back to the reference}};

\begin{scope}[shift={(0,-7.1)}]

  \def\pw{3.7}   
  \def\ph{2.6}   

  \begin{scope}[shift={(0.7,0)}]
    \draw[rounded corners=4pt, draw=scaleBorder, fill=scalePanel, thick]
          (0,0) rectangle (\pw,\ph);
    \foreach \i in {0,1,2,3,4}{
      \foreach \j in {0,1,2}{
        \pgfmathsetmacro\xx{0.55+\i*0.65}
        \pgfmathsetmacro\yy{0.55+\j*0.75}
        \pgfmathsetmacro\len{(\i==2)?0.62:0.42}
        \draw[abBorder,thick,->] (\xx,\yy+\len/2) -- (\xx,\yy-\len/2);
      }
    }
    \node[title,abText] at (\pw/2,\ph+0.45) {scale field $s(y;\nu)$};
    \node[font=\footnotesize] at (\pw/2,-0.5)  {$s(y)=\alpha(y)\!\cdot\!\nu+\beta(y)\!\cdot\!\nu^2$};
    \node[capt] at (\pw/2,-0.95) {controls local stretch / compression};
  \end{scope}

  \begin{scope}[shift={(5.65,0)}]
    \draw[rounded corners=4pt, draw=shiftBorder, fill=shiftPanel, thick]
          (0,0) rectangle (\pw,\ph);
    \foreach \i in {0,1,2,3,4}{
      \foreach \j in {0,1,2}{
        \pgfmathsetmacro\xx{0.55+\i*0.65}
        \pgfmathsetmacro\yy{0.55+\j*0.75}
        \pgfmathsetmacro\ang{22+\i*2-\j*3}
        \draw[gdBorder,thick,->] ($(\xx,\yy)-(\ang:0.275)$) -- ++(\ang:0.55);
      }
    }
    \node[title,gdText] at (\pw/2,\ph+0.45) {shift field $t(y;\nu)$};
    \node[font=\footnotesize] at (\pw/2,-0.5)  {$t(y)=\gamma(y)\!\cdot\!\nu+\delta(y)\!\cdot\!\nu^2$};
    \node[capt] at (\pw/2,-0.95) {controls local translation};
  \end{scope}

  \begin{scope}[shift={(10.6,0)}]
    \draw[rounded corners=4pt, draw=warpBorder, fill=warpPanel, thick]
          (0,0) rectangle (\pw,\ph);
    \coordinate (cw) at (\pw/2,\ph/2);
    \foreach \r in {0.45,0.85,1.2}
       \draw[nomPurple,thick] (cw) circle (\r);
    \def\wx{1.90}\def\wy{1.00}
    \foreach \r in {0.55,0.95,1.35}
       \draw[defOrange,thick,dashed,smooth]
          plot[domain=0:360,samples=84,variable=\t]
          ({\wx + \r*cos(\t)*0.88 + 0.45*(\r*sin(\t)*0.58)},
           {\wy + \r*sin(\t)*0.58 + 0.5*(\r*cos(\t)*0.88)^2});
    \node[title] at (\pw/2,\ph+0.45) {combined warp};
    \node[font=\footnotesize] at (\pw/2,-0.5)  {$y_{\mathrm{nom}}=y\cdot\exp(s)+t$};
    \node[capt] at (\pw/2,-0.95) {affine in $y$, quadratic in $\nu$};
  \end{scope}

  \begin{scope}[on background layer]
    \node[rounded corners=8pt, draw=bottomBorder, fill=bottomFill, thick,
          fit={(0.0,-1.25) (15.05,\ph+1.05)}] {};
  \end{scope}
  \node[title] at (7.5,\ph+1.45)
        {The coefficient fields define the correction $T_\nu$ at each location};

\end{scope}

\end{tikzpicture}

%% file: figs/fnf_factorizable_tikz.tex
\begin{tikzpicture}[
    node distance=3cm,
    font=\large\sffamily,
    layer/.style={rectangle, draw=black!80, fill=blue!10, very thick, minimum width=4cm, minimum height=2.5cm, rounded corners},
    neuron/.style={circle, draw=black!80, fill=white, thick, minimum size=1cm},
    op/.style={circle, draw=black!80, fill=green!10, thick, minimum size=1.3cm},
    param/.style={rectangle, draw=black!80, fill=red!10, thick, minimum width=1.8cm, minimum height=1.3cm},
    data/.style={rectangle, draw=black!80, fill=yellow!10, thick, minimum width=2.5cm, minimum height=1.3cm},
    connection/.style={->, -{Stealth[length=4mm]}, thick, color=black!70},
    box/.style={draw=gray, dashed, rounded corners, inner sep=0.7cm}
]

\node[data, font={\huge\sffamily}] (inputs) {Inputs $y, x$};

\node[layer, above right=0.8cm and 3.5cm of inputs, align=center, font=\Large] (psi1) {Net $\Psi^{(1)}$\\(Masked MLP)};
\node[layer, below right=0.8cm and 3.5cm of inputs, align=center, font=\Large] (psiK) {Net $\Psi^{(K)}$\\(Masked MLP)};
\node at ($(psi1)!0.5!(psiK)$) (dots_mid) {$\vdots$};

\draw[connection] (inputs.east) -- ++(1.5,0) |- (psi1);
\draw[connection] (inputs.east) -- ++(1.5,0) |- (psiK);

\node[right=1.5cm of psi1, align=left, font={\Large\sffamily}] (coeffs1) {$\alpha^{(1)}, \beta^{(1)}$\\$\gamma^{(1)}, \delta^{(1)}$};
\node[right=1.5cm of psiK, align=left, font={\Large\sffamily}] (coeffsK) {$\alpha^{(K)}, \beta^{(K)}$\\$\gamma^{(K)}, \delta^{(K)}$};

\draw[connection] (psi1) -- (coeffs1);
\draw[connection] (psiK) -- (coeffsK);

\node[op, right=1.8cm of coeffs1, font=\Large] (mix1) {$\times, +$};
\node[op, right=1.8cm of coeffsK, font=\Large] (mixK) {$\times, +$};

\draw[connection] (coeffs1) -- (mix1);
\draw[connection] (coeffsK) -- (mixK);

\node[param, above=1.2cm of mix1, font=\Large] (nu1) {$\nu_1$};
\node[param, below=1.2cm of mixK, font=\Large] (nuK) {$\nu_K$};

\draw[connection, dashed, color=red!70] (nu1) -- (mix1) node[midway, right, font=\large, align=left] {Linear/\\Quad};
\draw[connection, dashed, color=red!70] (nuK) -- (mixK) node[midway, right, font=\large, align=left] {Linear/\\Quad};

\node[right=0.8cm of mix1, font=\Large] (contrib1) {$\Delta s^{(1)}, \Delta t^{(1)}$};
\node[right=0.8cm of mixK, font=\Large] (contribK) {$\Delta s^{(K)}, \Delta t^{(K)}$};

\path (contrib1) -- (contribK) coordinate[midway] (center_right);
\node[op, minimum size=1.8cm, right=2cm of center_right, font=\huge] (sum) {$\sum$};

\draw[connection] (contrib1) -| (sum);
\draw[connection] (contribK) -| (sum);

\node[right=1.5cm of sum, align=center, font={\Large\sffamily}] (final_params) {Total Scale $s$\\Total Shift $t$};
\draw[connection] (sum) -- (final_params);

\node[op, below=2cm of final_params, fill=blue!20, font=\Large] (transform) {Transform};
\node[data, right=2cm of transform, font={\huge\sffamily}] (output) {$y_{\text{nom}}$};

\draw[connection] (final_params) -- (transform);

\draw[connection, rounded corners=8pt] (inputs.south) -- ++(0, -7cm) -| (transform.south)
    node[pos=0.1, yshift=-0.4cm, right, font=\Large] {Input $y$}; 

\draw[connection] (transform) -- (output);

\node[below=0.5cm of output, align=left, font={\Large\sffamily}] {
    $y_{\text{nom}} = y \cdot e^s + t$\\[0.6em]
    $s = \sum (\alpha \nu + \beta \nu^2)$\\[0.6em]
    $t = \sum (\gamma \nu + \delta \nu^2)$
};

\begin{pgfonlayer}{background}
    \node[box, fit=(nu1) (psi1) (mix1) (contrib1), label=above:Systematic 1] {};
    \node[box, fit=(nuK) (psiK) (mixK) (contribK), label=below:Systematic K] {};
\end{pgfonlayer}

\end{tikzpicture}